\documentclass[sigconf,nonacm,authorversion]{acmart}

\AtBeginDocument{%
  \providecommand\BibTeX{{%
    \normalfont B\kern-0.5em{\scshape i\kern-0.25em b}\kern-0.8em\TeX}}}

\acmConference[ICONS 2020]{International Conference on
Neuromorphic Systems}{July 28--30, 2020}{Virtual Conference}

\begin{document}

\title[Coarse scale models of spiking neural networks]{Coarse scale representation of spiking neural networks: backpropagation through spikes and application to neuromorphic hardware}

\author{Angel Yanguas-Gil}
\affiliation{%
  \institution{Applied Materials Division, Argonne National Laboratory}
  \city{Lemont}
  \state{Illinois}
  \postcode{60439}
}
\email{ayg@anl.gov}
\orcid{1234-5678-9012}


\begin{abstract}
In this work we explore recurrent representations of 
leaky integrate and fire neurons operating at a timescale equal
to their absolute refractory period. Our coarse time scale approximation is obtained using a probability distribution function for spike arrivals that is homogeneously distributed over this time interval. This leads to a discrete representation that exhibits the same dynamics as the continuous model, enabling
efficient large scale simulations and backpropagation through the recurrent implementation.
We use this approach to explore the training of deep spiking neural networks including convolutional, all-to-all connectivity, and maxpool layers directly in Pytorch.
We found that the recurrent
model leads to high classification accuracy using just 4-long spike trains during
training. We also observed a good transfer back to continuous implementations of leaky integrate
and fire neurons. Finally, we applied this approach to some of the standard control problems 
as a first step to explore reinforcement learning using neuromorphic chips.

\end{abstract}

\ccsdesc[500]{Computing methodologies~Neural networks}
\ccsdesc[300]{Computing methodologies~Machine learning}
\keywords{spiking neurons, neural networks, neuromorphic
computing, machine learning}

\maketitle

\section{Introduction}

Spiking neurons have long been the subject of intense study due to their central role in the central nervous system. Recently,
there has been a 
renewed interest in the subject in the
context of artificial intelligence and neuromorphic hardware,
where spiking neurons offer the promise of power-efficient
computing architectures capable of tackling problems 
that are hard to solve using conventional
computational approaches.

One  of the fundamental challenges of using spiking neurons
for computations or machine learning applications is how
to adapt the stochastic gradient descent methods at the
core of training algorithms for artificial neural networks to 
their spiking counterparts. The motivation for doing so  is
twofold: first, it provides
a way of implementing machine learning algorithms in neuromorphic
hardware. Second, it facilitates the use of existing machine
learning tools 
to explore the computational capabilities of biological
neural networks.

Not surprisingly, this is a problem that has
been repeatedly tackled in the literature.
The proposed approaches can be split into two broad categories: 
one type of approach has focused on how to efficiently transfer
trained networks from artificial neural networks to their
spiking counterparts. The second type of approach
explores heuristic approximations for gradients
or algorithms in spiking networks themselves.
Both have yielded promising results.\cite{
Bohte2002, hunsberger2015spiking, diehl2016conversion,
Lee2016, Sengupta2019, Lee2020}

In this work, we explore the connection between
piece-wise differentiable continuous models of spiking neurons
and recurrent neural networks. Our motivation is that,
from
a machine learning perspective, modeling spiking
neurons is very inefficient: the time scale used
in the discretization of the differential equations controlling the neuron dynamics is typically much
smaller than the timescales at which information
propagates through the network. If we find coarse-scale
models that evolve at a faster pace, we can 1) efficiently implement larger networks 2) minimize the number of
timesteps required when training spiking neural networks
using stochastic gradient descent methods.

In particular, we exploit the presence of
an absolute refractory period defining
the largest
timescale at which a neuron can spike \emph{at most once}. 
 A model that
is capable of evolving at this pace while reproducing the underlying
dynamics of the spiking neuron would provide a very efficient
implementation. The existence
of such system is however by no means guaranteed:  there is no one-to-one
correlation between a system of differential equations and a 
discrete implementation with arbitrarily large time steps.

In this work, we therefore seek to understand how
we can create coarse-scale approximations of spiking systems that allow us to map spiking models into recurrent systems that evolve at
time scales of the order of the absolute refractory period
of neurons. Our approach is based on managing the loss of information
on the exact timings of the system at timescales smaller than
our evolution step by transforming the differential equations
of leaky integrate and fire (LIF) neurons into a probabilistic models. By
 making fundamental assumptions about the distributions characterizing spike arrival and spike generation we can derive
different coarse-scale models. 

This paper is structured as follows: we first
derive three different coarse-scale models from a standard leaky
integrate and fire equations. We then benchmark these models
against the LIF neurons by exploring sparsely connected
spiking neurons. After demonstrating the equivalence of
some of these models, we implement stochastic gradient
descent methods and explore the training of shallow and deep spiking
neural networks using this approach. Finally, we explore the
use of spiking networks for control tasks as a first step for
their implementation in neuromorphic hardware.

\section{Model}
\label{sec:model}

This work uses the leaky integrate and fire
neuron as a model system for spiking neurons.  In a LIF model,
the membrane potential $v_i$ of the neuron
is given by:
\begin{equation}
\label{eq_lif}
\tau \frac{d v_i}{dt} = - v_i + \tau \sum_j w_{ij} \delta(t-t_{ij}) + \tau u_\mathrm{ext}^{(i)}
\end{equation}
Here spikes are treated as Dirac's delta impulses, $w_{ij}$ are the synaptic weights,
 and $\tau$ is the leakage time of the membrane potential of the neuron, and $u_\mathrm{ext}^{(i)}$ is a term comprising non-spiking external currents. 
 Each neuron is subject to the spike firing condition whenever the membrane potential reaches a threshold value $v^{(i)}_0$:
\begin{equation}
v_i(t) = v^{(i)}_0 \Rightarrow v_i(t + t') = 0 \hspace{0.5cm} \forall \hspace{0.5cm} t' < \tau_r
\end{equation}
For simplicity, we have chosen a reset potential $v_\mathrm{reset}=0$, but generalizations to other values are straightforward.
The key parameter of the model is the absolute refractory period
$\tau_r$, defining the time after a spike during which the membrane potential is not receptive towards incoming input. This absolute refractory
period provides the coarsest timescale during which each neurons spikes at most once. It is therefore the natural
timescale that we can use to build an efficient coarse-scale representation of this type of neurons.

Eq. \ref{eq_lif} can be integrated over a timescale $\Delta t$ to obtain:
\begin{equation}
\label{eq_disc1}
\begin{array}{rcl}
v_i(t+\Delta t) & = &  v_i(t) e^{-\Delta t/ \tau}  +   \tau u_\mathrm{ext}^{(i)} \left(1- e^{-\Delta t / \tau} \right) \\
& & +  \sum_{j: t \le t_{ij} < t + \Delta t} w_{ji} e^{-(t+\Delta t-t_{ij})/\tau}
\end{array}
\end{equation}

where the sum is extended to all spikes received within the $t$ and $t  + \Delta t$ interval. Here we have assumed that  $u_\mathrm{ext}^{(i)}$ change
slowly during that interval.

As we move to a coarse scale representation we lose information on the exact firing times $\Delta t$. 
Therefore, we
assume that the actual spike arrival times $t_{ij}$ are homogeneously distributed between $t$ and  $t+\Delta t$ with a probability
density $1/\Delta t$. This leads to the following averaged contribution of incoming spikes to Eq. \ref{eq_disc1}:
 \begin{equation}
\langle e^{-(t+\Delta t-t_{ij})/\tau} \rangle = \frac{1}{\Delta t} \int_t^{t+\Delta t}
 e^{-(t+\Delta t -t_{ij})/\tau} dt_{ij}
= \frac{\tau}{\Delta t}\left( 1 - e^{-\Delta t /\tau} \right)
\end{equation}
For the specific case of $\Delta t = \tau_r$, and defining $s_i=0,1$ depending on whether the neuron spikes in each
interval, we have that:
\begin{equation}
\label{eq_nofire}
v_i(n) = v_i(n-1) e^{-\tau_r/\tau} + \xi_i(n)\left(1- e^{-\tau_r/\tau}\right) 
\end{equation}
and
\begin{equation}
\label{eq_xi}
\xi_i(n)= \tau u_\mathrm{ext}^{(i)} (n)+ \frac{\tau}{\tau_r}  \sum_j  w_{ij}  s_j(n-n_{ij})
\end{equation}
where
\begin{equation}
\label{eq_heavy}
s_i(n) = H\left(v_i(n)-v_0^{(i)}\right)
\end{equation}
where $H(\cdot)$ is the Heaviside or step function and $v_0^{(i)}$ is the firing threshold.

Eq. \ref{eq_nofire} assumes that the neuron did not fire in the prior interval. If a neuron spiked in the past interval
we need to account for the refractory period, during which the voltage is assumed to be clamped to its reset value. Depending 
on the assumption that we make about when a neuron spikes, we can define three different models:

\subsection{Model I}

Model I assumes
that the precise instant in which the neuron leaves its refractory period is uniformly distributed over $\Delta t = \tau_r$. This
means that the neuron will be receptive to external inputs only during a fraction $\delta t$ within
the interval $\Delta t$, so that:
\begin{equation}
v_i(t+\delta t) =  v_\mathrm{ext}^{(i)} \left(1- e^{-\delta t / \tau} \right)
+ \sum_{j: t \le t_{ij} < t + \delta t} w_{ji} e^{-(\delta_t-t_{ij})/\tau}
\end{equation}
we then have to average over $t_{ij}$ and $\delta t$ to obtain:
\begin{equation}
\label{eq_fire}
v_i(n) =  \xi_i(n)\left(1- \frac{\tau}{\Delta t}(1-e^{-\Delta t/\tau})\right)
\end{equation}

If we now particularize $\Delta t = \tau_r$, we have replaced the asynchronous leaky integrate and
fire model with a recurrent discrete time difference equations that neglects timescales smaller than 
the absolute refractory period of the networks. 

Eqs. \ref{eq_nofire} and \ref{eq_fire} can be concisely expressed as:
\begin{equation}
\label{eq_onerefract}
\begin{array}{rcl}
v_i(n) & = & (1-s_i(n-1)) \left[ v_i(n-1) e^{-\tau_r/\tau} + \xi_i(n)\left(1- e^{-\tau_r/\tau}\right)\right] \\
& & + s_i(n-1) \xi_i(n) \left(1- \frac{\tau}{\tau_r}(1-e^{-\tau_r /\tau})\right) \\
\end{array}
\end{equation}

\subsection{Model II}

Model  II considers that the neuron fires at the
end of the time interval, clamping the 
potential to its reset value during the next step.
Consequently, while neurons in 
Model I can spike during each interval, in Model II 
neurons can spike at most once every two intervals.

The resulting recurrent expression for this model is
therefore:
\begin{equation}
\label{eq_simpmodel}
v_i(n)  =  \left(1-s_i(n-1) \right) \left[ v_i(n-1) e^{-\tau_r/\tau} + \xi_i(n)\left(1- e^{-\tau_r/\tau}\right)\right] 
\end{equation}

\subsection{Model III}

Model III resets the potential but otherwise the system is receptive to spikes during the next interval after the
neuron spikes:
\begin{equation}
\label{eq_simpmode2l}
v_i(n)  =  \left(1-s_i(n-1) \right) v_i(n-1) e^{-\tau_r/\tau} + \xi_i(n)\left(1- e^{-\tau_r/\tau}\right)
\end{equation}

In all three cases, $s_i(n) = H\left(v_i(n)-v_0^{(i)}\right)$.

\section{Validation of the coarse-scale approximation}

To validate the accuracy of the three models, we have compared their performance with
their corresponding asynchronous leaky integrate and fire. We have considered a network of 1000 randomly connected neurons with a connectivity
density of 5\%. This type of networks has been well characterized in the literature, and it is characterized by a complex dynamics. It is
therefore the perfect model to explore the impact of the coarse
scale models.

\begin{figure*}[ht]
\centering
\includegraphics[width=15cm]{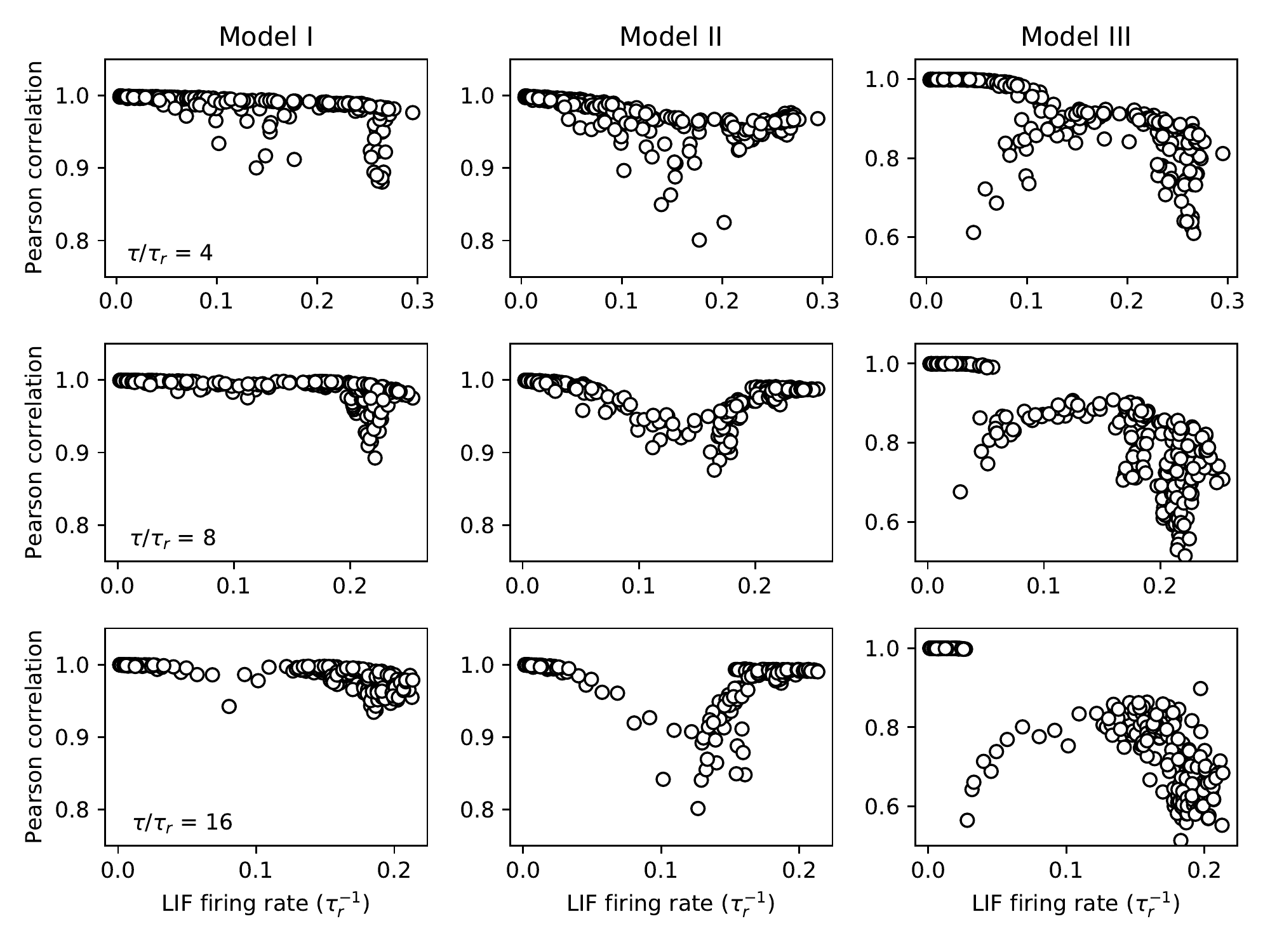}
\caption{\label{fig_corrfull} Pearson correlation between the firing rates (spikes per unit time) of the asynchronous LIF and the synchronous
representation for $\Delta t = \tau_r$ as a function of the average activity of the network. Results  are shown  for Models I, II, and III,
and three values of $\tau/\tau_r$.
The loss of information on the exact spiking times has only marginal effects for Models I and II the activity is low.}
\end{figure*}

Here we consider  the case in which neurons are excited with constant external inputs sampled from a Gaussian distribution. In Figure \ref{fig_corrfull} we show the Pearson correlation
between the average activities of the LIF models and the three discrete approximations. Each point corresponds to a different network randomly
instantiated and with randomly selected inputs. Results are shown for three values of $\tau/\tau_r$.

Despite the loss of information on the timing of spikes at timescales
shorter than the absolute refractory period, the results indicate a
good agreement between
the average activity of Model I and the LIF model.
Model II has also a reasonably good agreement.
However, Model III fails to reproduce the dynamics of
the recurrent LIF network (note the different  vertical scale for
the correlation in Figure \ref{fig_corrfull}). This emphasizes the importance
of managing the information about spike timings that is lost
when shifting to coarser time scales.

\section{\label{sec_stochastic} Stochastic gradient descent implementation}

After validating that Model I and Model II agree reasonably well with
the LIF model, the implementated stochastic gradient
descent methods in these two models.
Both Eq. \ref{eq_onerefract} and \ref{eq_simpmodel} allow
us to represent
a spiking network as a recurrent layer of the output spike $\mathbf{s}$ and
the membrane potential $\mathbf{v}$,
so  that:
\begin{equation}
    \label{eq_layer}
    \mathbf{s}(t),  \mathbf{v}(t) = f\left(\mathbf{\xi}(t); \mathbf{s}(t-1),  \mathbf{v}(t-1)  \right)
\end{equation}
where  $\mathbf{s}$ and $\mathbf{v}$ are related  through:
\begin{equation}
\label{eq_hard}
    \mathbf{s}=H\left(\mathbf{v}-\mathbf{v}_0\right)
\end{equation}

In order to enable stochastic gradient descent methods, we
calculate $\mathbf{s}$ using Eq. \ref{eq_hard} in the forward direction,
whereas in the backward direction gradients are calculated using
a differentiable approximation. Here, we have used the logistic function $\sigma(\cdot)$, so that:
\begin{equation}
\label{eq_backward}
\mathbf{s}_\mathrm{back} = \sigma\left(\beta(\mathbf{v}-\mathbf{v}_0\right))
\end{equation}
Here $\beta$ is a regularization parameter that determines the steepness of the approximation.
n this work,
we implemented these models directly in \verb|Pytorch|. 
Details on the implementation are included in the Appendix, and the code can be found
online at: \verb|https://github.com/anglyan/spikingtorch|.

While  Eq. \ref{eq_layer} allows us to train networks to match
specific spike trains, here we have considered the total activity
of the output neurons:
\begin{equation}
\mathbf{a} = \sum_t \mathbf{s}(t)
\end{equation}
This  would in principle allow us to consider two
different approaches: we can try to  match a specific number of
spikes $N_{out}$, or we can use a cross-entropy cross-section that
tries to maximize the activity of specific neurons. Our experience
shows that cost functions such as MSE that take into account
the number of output spikes perform more consistently than the cross-entropy method.

In our recurrent implementation of spiking neurons, inputs
and outputs are codified as a time sequence of length $N_{sp}$. 
We have considered the following
four type of encodings:
\begin{enumerate}

\item \emph{random spike train}. Input values are codified as 
a Bernouilli distribution, the discrete equivalent of Poisson
spike trains.

\item \emph{Periodic spike  train}. A periodic train
of spikes spaced by an interval that is inversely related
to the intensity of the input.

\item \emph{Single spike delay encoding}. In this case the input is codified as a single spike whose delay with respect to a
common epoch decreases with the input intensity.

\item \emph{Constant analog input}. The input is codified as constant input signal.

\end{enumerate}

\section{Results}

\subsection{\label{sec:multilayer} Shallow and multilayer spiking networks}

\begin{table*}
  \caption{Classification accuracy of trained
  spiking networks based on Model I for 8-long 
  Bernouilli input spike trains. Two non-spiking
  benchmarks are shown for comparison: one
  where inner spiking layers are replaced with rectified
  linear units and the output with a softmax,
  and a second case where all spiking layers were replaced
  with sigmoid activation functions.
  All cases were trained for
  15 epochs}
  \label{tab:multi}
  \begin{tabular}{lcc|cccc}
    \toprule
     &  \multicolumn{2}{c}{Spiking} & 
        \multicolumn{2}{c}{ReLU+SoftMax} &
        \multicolumn{2}{c}{Sigmoid}\\
     \midrule
    Network &  MNIST  &  FMNIST &  MNIST  &  FMNIST &  MNIST  &  FMNIST  \\
    \midrule
    Shallow & 91.0 & 81.4 & 92.9 & 84.1 & 92.0 & 83.7 \\
    In/Lin/Sp/Lin/Sp & 95.6 & 84.7 & 96.8 & 86.6 & 95.0 & 86.4\\
    In/Conv/Sp/Lin/Sp & 96.4 & 84.9 & 97.6 & 87.3 & 97.0 & 86.6 \\
    In/Conv/Sp/Conv/Sp/Lin/Sp & 97.9 & 85.7 & 98.4 & 87.8 & 97.8 & 86.3 \\
  \bottomrule
\end{tabular}
\end{table*}

In order
to test the proposed approach we first tested 
its performance in classification tasks involving
the MNIST and Fashion MNIST datasets.\cite{Lecun1998,xiao2017fashionmnist}.
We considered the following networks:

\begin{itemize}

    \item \textbf{Shallow network}: input spikes are densely connected
    to the output neurons with no bias.
    
    \item \textbf{In/Lin/Sp/Lin/Sp}: a densely connected network with
    one hidden layer of spiking neurons. The results shown
    in this work have been obtained using 30 hidden
    neurons.
    
    \item \textbf{In/Conv/Sp/Lin/Sp}: a simple convolutional network with
    a 2D convolutional layer composed of a 5x5 kernel with a 
    stride and padding of 2 outputting to four independent
    channels. This results on 784 spiking neurons that are
    then fully connected to the spiking output layer through a 
    linear layer with no bias.
    
    \item \textbf{In/Conv/Sp/Conv/Sp/Lin/Sp}: a convolutional network
    with two 2D convolutional layers, each with a 5x5 kernel. The
    first layer, with the same configuration as the single layer case,
    has a stride and padding of 2 outputting to four independent
    channels. The second convolutional layer has a stride of 1 and
    6 output channels. The output
    is then fed to a spiking layer that is fully connected to the
    spiking output layer with no bias.
    
\end{itemize}

The additional case of a spiking LeNet5 is analyzed in Section \ref{sec:lenet}.

These networks, implemented using Model I,
were trained against the MNIST and Fashion MNIST (FMNIST)
datasets, resulting  on the classification accuracy values
shown in Table \ref{tab:multi}.
These results were obtained for an 8-long spike trains trained for 15 epochs with a MSE cost function
and a desired activation pattern of the output
layers of 4 spikes for the correct neuron.
The threshold voltage was kept constant and
equal to $v_0^i=1$.  In Table \ref{tab:multi} we also
show the comparison with two separate benchmarks:
in the first benchmark, the spiking layers were
replaced with rectified linear units except for
the output layer, where a softmax function is
used. This first benchmark was trained using a
cross-entropy loss function. The second benchmark
replaced all spiking neurons with sigmoid layers,
and the training is carried out using an MSE
cost function, the same as the spiking case. A third benchmark in which all
neurons are replaced with rectified linear units was briefly considered, but
it failed to achieve good classification accuracies in FMNIST.

In all cases, the stochastic gradient descent method was
able to converge to an accuracy comparable to that of the non-spiking counterpart, demonstrating  the
ability to carry out backpropagation both through the
depth of the network and through spike trains.
Access to efficient implementations
of convolutional layers and GPU compatibility greatly aided  the
computational efficiency of the training process, resulting on a significant
acceleration compared to CPU-based approaches. Moreover, a comparison between
the two benchmarks shows that sigmoidal activation functions lead to lower accuracies
compared to the ReLU benchmark. This may suggest that the bounded nature of the output
of spiking neurons, together with smaller gradients far from the firing threshold, could
account for the small dip in accuracy with respect to the ReLU benchmark.

\subsection{Analysis of the shallow network case}

We have used the shallow network case to look
more in depth to some of the particularities
of the recurrent models used in this work.

First, we have looked at the transfer
of trained recurrent networks to the conventional
LIF model represented by the differential Eq. \ref{eq_lif}. This is something that is relevant,
for instance, if we want to transfer
trained networks to neuromorphic hardware or
to a software model providing a exact implementation
of LIF neurons. 

\begin{figure}[ht]
\centering
\includegraphics[width=8.5cm]{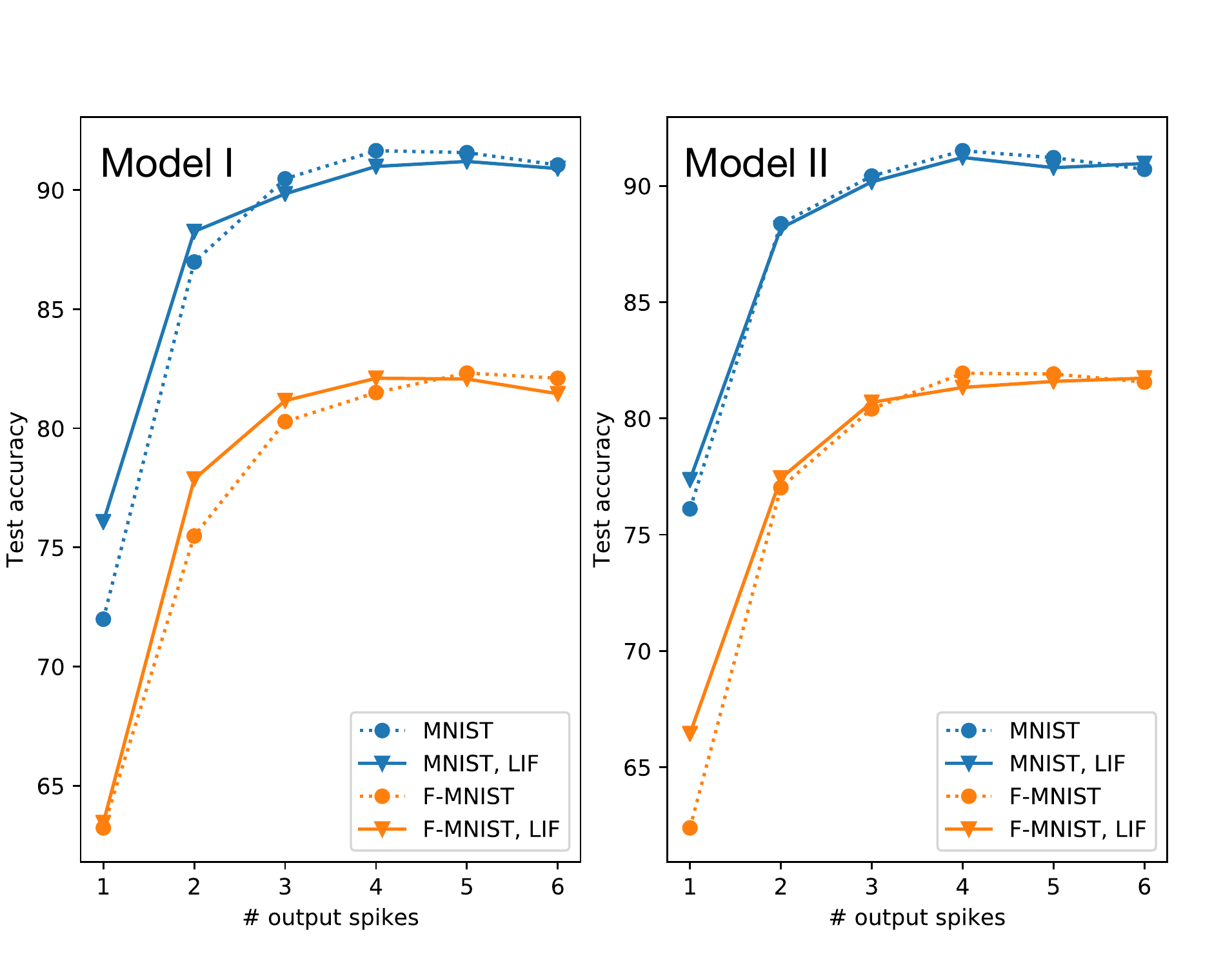}
\caption{\label{fig_model} accuracy of a shallow
spiking network implementing Model I (left)
and Model II (right) in MNIST and Fashion MNIST
classification tasks. The accuracy of the coarse-scale model
and that of the full LIF model on the trained network is shown as
a function of the number of spikes used in the training process.}
\end{figure}

In Figure \ref{fig_model} we
show the accuracy achieved by a simple shallow spiking
network of Models I and II, and we compare it
with the resulting efficiency when the same
network is transferred back to the fine-scale LIF
model where spikes  are treated as Dirac's delta
functions.  We observe an excellent
correlation between the classification accuracy obtained
in the coarse scale and in the LIF models.

Then we explored the impact that the number of
target output neurons spikes $N_{out}$ used during
training had on the network
accuracy. As shown in Figure \ref{fig_model},
there is a minimum number of output spikes
required in order to achieve the maximum
accuracy. For 8-long spike trains we 
need to require at least four output
spikes in the correct output neuron for the algorithm
to effectively carry out the classification task.
This is observed for both Model I and Model II.

Finally, we explored the impact that
the slope parameter $\beta$ used in the
regularization of the Heaviside function (Eq. \ref{eq_backward} 
has on 
the ability to classify both datasets.
As shown in Table \ref{tab:beta}, there is a small
dependence with $\beta$, with both classification
accuracy and the agreement between the coarse-scale
and the LIF model being slightly better for
$\beta \ge 3$.

\begin{table}
  \caption{Impact of the slope parameter $\beta$ controlling spike regularization on the
  classificaction accuracy of both MNIST
  and Fashion MNIST in a shallow network}
  \label{tab:beta}
  \begin{tabular}{c|ccccc}
    \toprule
     & \multicolumn{5}{c}{Value of $\beta$}  \\

    \midrule
      & 1 & 2 & 3 & 4 & 5  \\
    \midrule
    MNIST & 90.6 & 91.0 & 91.0 & 90.9 & 90.7  \\
    FMNIST & 79.7 & 80.7 & 81.4 & 81.7 & 81.8  \\
  \bottomrule
\end{tabular}
\end{table}

\subsection{\label{sec:lenet} Spiking LeNet5}

To deepen our understanding of the design principles
of deep spiking networks we have focused on a spiking
version of the LeNet5 network.\cite{Lecun1998} The spiking version
of this network maintains the convolutional and 
maxpool layers of the non-spiking version and replaces
the five non-linear layers with spiking neuron 
counterparts.

Here we have compared two different encodings: a random
spike train, based on a Bernouilli distribution
with an average equal to the pixel intensity,
and a periodic spike train. For the periodic
spike train cases we added a sixth layer of spiking
neurons where the pixel intensities, normalized from 0 to 1,
are transformed into a periodic train of spikes, mimicking
the behavior of sensory neurons. The
rest of the network remains the same.
We have also compared the case where biases are omitted
from the three densely connected layers in LeNet5. This
is consistent with the presence of a constant firing
threshold for all the neurons in the network.

In Table \ref{tab:lenet} we show a comparison between
these four cases for the MNIST and F-MNIST datasets. 
We have considered the same two benchmarks used
in Section \ref{sec:multilayer}. We report top-one accuracy values in all cases.
The results obtained are
remarkable insensitive towards the type of encoding.
This is somewhat unexpected given the dependence on 
accuracy with spike length reported in other methods,
which required up to 150 steps for the more complex
networks. Also, the presence of additional bias
in  the fully connected layers do not seem to greatly
impact the network's accuracy. It is feasible that some of
these differences could be fleshed out with different training
strategies or by training for a considerable larger number of
epochs. 

\begin{table}
  \caption{Classification accuracy of spiking LeNet
  network for 8-long spike trains. All cases were trained
  during 15 epochs using the same training conditions.}
  \label{tab:lenet}
  \begin{tabular}{llcc}
    \toprule
    Encoding & Bias &  MNIST  &  FMNIST \\
    \midrule
    Poisson & yes & 98.8 & 87.5 \\
    Periodic & yes & 98.6 & 88.0 \\
    Poisson & no & 98.6 & 87.5 \\
    Periodic & no & 98.6 & 87.4 \\
    \midrule
    Benchmarks & & & \\
    \midrule
    ReLU + SoftMax & yes & 98.8 & 90.3 \\
    ReLU + SoftMax & no & 99.0 & 89.6 \\
    Sigmoid & yes & 98.8 & 89.5\\
    Sigmoid & no & 98.8 & 88.4 \\
 
  \bottomrule
\end{tabular}
\end{table}

Finally, in order to evaluate the impact of
the length of the spike trains used during training, we performed
an experiment where networks trained with
random spike trains of a given length where
then tested using spike trains of varying
lengths. The results, summarized in Table \ref{tab:transfer},
show how networks trained with as few as 4-long spike
trains can be successfully transferred during inference
to inputs with longer 
spike trains, resulting on higher classification accuracy.
Still, it is important to note that the difference in
accuracy for short and long spike trains is small, ranging
between 0.5 and 1\%.

\begin{table*}
  \caption{Impact of spike train length
  during training and testing in spiking LeNet's accuracy on
  MNIST. Values
  are the average of ten independent tests, with the standard
  deviation in the second decimal place.}
  \label{tab:transfer}
  \begin{tabular}{c|ccccccc}
    \toprule
    Input length & \multicolumn{7}{c}{Testing}  \\

    \midrule
    Training & 4 & 6  & 8 & 12 & 16 & 32 & 48 \\
    \midrule
    4 & 98.52 & 98.82 & 98.91 & 98.98 & 99.00 & 99.02 & 99.00 \\
    6 & 98.48 & 98.82 & 98.90 & 98.96 & 98.99 & 99.03 & 99.03 \\
    8 & 98.43 & 98.81 & 98.90 & 98.97 & 98.97 & 99.01 & 99.00 \\
    12 & 98.21 & 98.70 & 98.85 & 98.95 & 99.00 & 99.07 & 99.10 \\
  \bottomrule
\end{tabular}
\end{table*}

\subsection{Application to neuromorphic hardware and
control tasks}

As shown in a previous work, our implementation of the LIF model
provides an excellent agreement with the fixed-point implementation
in Loihi.\cite{Loihi,aygspace} Therefore, we can directly use these models to train
and port spiking neural network into neuromorphic
hardware. We have
also applied this model to explore the optimization of neuromorphic
architectures implemented using cross-point arrays.\cite{aygmem}

In order to evaluate the feasibility of training
LIF neurons directly using backpropagation for
control tasks, we have explored their application
to standard reinforcement learning benchmarks
such as Cartpole.\cite{Barto} We have considered a 
spiking neural network in which the four-dimensional
inputs from the environment are passed through
a densely connected linear layer into a hidden layer
of spiking neurons, which are in turn densely
connected to the two output spiking neurons. 

The output layer accumulates the number of spikes $\mathbf{n}_s$
over
a finite number of steps $N_{sp}$, and transforms it into
a value that can be directly used to calculate the
response probabilities:
\begin{equation}
\mathbf{y} = \sigma\left[\alpha \left(\mathbf{n}_s - \frac{N_{sp}}{2}\right)\right]
\end{equation}
where $\alpha$ is a scaling parameter used to
ensure that $\mathbf{y}$ covers a sufficiently large
fraction of the $[0,1]$ interval.

We have used a naive implementation of the cross-entropy algorithm to dynamically learn the optimal policy by
training the network using the upper 30\% percentile of
episodes in each batch. For a hidden layer of 64 neurons we
can achieve an average of 199.3 steps over 100 consecutive
episodes with 8-long spike trains
using fewer than 50 batches of 32
episodes.

\section{Conclusions}

In this work we have explored recurrent representations of 
leaky integrate and fire neurons operating at a timescale equal
to their absolute refractory period. This leads to highly efficient implementations that present identical dynamics, which provides unique opportunities for large scale simulations and
the efficient emulation of existing neuromorphic hardware implementing leaky integrate and fire neurons.

The exploration of coarse-scale representation of leaky integrate and fire networks leads to a straightforward implementation of 
stochastic gradient descent method in spiking neurons without
the need of approximating the gradients beyond the regularization
of a step-function during the backward step. We have
explored the application of this methodology to
networks with up to six layers of spiking neurons, and
we have shown that training a spiking LeNet5 network
with 4-long spike trains is enough to achieve 99\%
accuracy in the MNIST task. Likewise, we have demonstrated
that 8-long spike trains are enough to train a spiking
network on the Cartpole task using the same backpropagation
methods.

The approach outlined in this work provides a straightforward entry point to explore the potential of spiking networks using
conventional machine learning frameworks. 

\begin{acks}
This research is based upon work supported by  Laboratory Directed Research and Development (LDRD) funding from Argonne National Laboratory, provided
by the Director, Office of Science, of the U.S. Department of Energy under Contract No. DE-AC02-06CH11357.
\end{acks}

\bibliographystyle{ACM-Reference-Format}
\bibliography{sample-base}

\appendix

\section{Implementation details}

The implementation of stochastic gradient descent methods
using the coarse scale representation derived in this work
requires the propagation of gradients over Heaviside functions using
a differentiable approximation.
The implementation of this function in a machine learning
framework such as \verb|Pytorch| is straightforward:

\begin{verbatim}

class HardSoft(torch.autograd.Function):

    @staticmethod
    def forward(ctx, input, constant):
       ctx.constant = constant
        output = torch.sigmoid(constant*input)
        ctx.save_for_backward(output)
        return H(input)

    @staticmethod
    def backward(ctx, grad_output):
        out, = ctx.saved_tensors
        return grad_output * out*(1-out)*ctx.constant, None

def H(x):
    return 0.5*(torch.sign(x)+1)
    
\end{verbatim}

The rest of the algorithms can be  implemented using the standard
methods as described in \verb|Pytorch|'s documentation. The code used to compute
some of the examples in this work
can be found in the following website:
\verb|https://github.com/anglyan/spikingtorch|.

\end{document}